\documentclass[sigconf]{aamas} 

\usepackage{balance} 
\usepackage{graphicx}
\usepackage{xcolor}
\usepackage{float}
\usepackage{color}
\usepackage{amsmath}
\usepackage{algorithm}
\usepackage{algpseudocode}
\usepackage{caption}
\usepackage[normalem]{ulem}
\useunder{\uline}{\ul}{}
\usepackage[section]{placeins}
\usepackage[export]{adjustbox}
\usepackage{subfigure}

\algnewcommand{\LeftComment}[1]{\Statex \(\triangleright\) #1}
\DeclareMathOperator*{\argmax}{arg\,max}
\algnewcommand{\OR}{\algorithmicor}

\setcopyright{none}
\acmConference[ALA '22]{Proc.\@ of the Adaptive and Learning Agents Workshop (ALA 2022)}
{May 9-10, 2022}{Online, \url{https://ala2022.github.io/}}{Cruz, Hayes, da Silva, Santos (eds.)}
\copyrightyear{2022}
\acmYear{2022}
\acmDOI{}
\acmPrice{}
\acmISBN{}
\acmSubmissionID{43}

\title[paper-title]{Methodical Advice Collection and Reuse \\in Deep Reinforcement Learning}

\author{Sahir}
\affiliation{
  \institution{Alberta Machine Intelligence Institute (Amii) \& \\University of Alberta}
  \city{Edmonton}
  \state{Alberta, Canada}}
\email{sahir1@ualberta.ca}

\author{Ercüment İlhan}
\affiliation{
  \institution{Queen Mary University of London}
  \city{London}
  \state{England, United Kingdom}}
\email{e.ilhan@qmul.ac.uk}

\author{Srijita Das}
\affiliation{
  \institution{University of Alberta}
  \city{Edmonton}
  \state{Alberta, Canada}}
\email{srijita1@ualberta.ca}

\author{Matthew E. Taylor}
\affiliation{
  \institution{University of Alberta \& \\
  Alberta Machine Intelligence Institute (Amii)}
  \city{Edmonton}
  \state{Alberta, Canada}}
\email{matthew.e.taylor@ualberta.ca}

\begin{abstract}
Reinforcement learning (RL) has shown great success in solving many challenging tasks via use of deep neural networks. Although using deep learning for RL brings immense representational power, it also causes a well-known sample-inefficiency problem. This means that the algorithms are data-hungry and require millions of training samples to converge to an adequate policy. One way to combat this issue is to use action advising in a teacher-student framework, where a knowledgeable teacher provides action advice to help the student. 
This work considers how to better leverage uncertainties about when a student should ask for advice and if the student can model the teacher to ask for less advice. The student could decide to ask for advice when it is uncertain or when both it and its model of the teacher are uncertain. 
In addition to this investigation, this paper introduces a new method to compute uncertainty for a deep RL agent using a secondary neural network. Our empirical results show that using dual uncertainties to drive advice collection and reuse may improve learning performance across several Atari games.

\end{abstract}

\keywords{Reinforcement Learning, Action Advising, Teacher-Student Framework}

\newcommand{\BibTeX}{\rm B\kern-.05em{\sc i\kern-.025em b}\kern-.08em\TeX}

\begin{document}


\pagestyle{fancy}
\fancyhead{}


\maketitle 


\section{Introduction}

Deep RL has shown promising results in many challenging problems, ranging from the game of Go~\cite{silver2016mastering} to  Atari video games~\cite{mnih2015human}. While these problems have opened up avenues for applying RL research to a wide variety of important problems like drug discovery~\cite{gottipati2021towered}, molecular optimization~\cite{zhou2019optimization} and healthcare~\cite{liu2019learning}, current Deep RL algorithms are \textit{sample-inefficient} and require millions of interactions with the environment to converge. This restricts deploying Deep RL systems in real-world applications where acquiring training samples incur cost.  


A number of well-known techniques like reward shaping~\cite{Ngetal99}, policy shaping~\cite{griffith2013policy} and imitation learning~\cite{pomerleau1991efficient} have been studied to address the sample-efficiency problem by using expert knowledge to better guide the agent. A popular paradigm for providing expert advice is a teacher-student framework~\cite{clouse1996integrating} where a teacher is an experienced agent, trained on the same or related task, who provides advice to a naive student agent to help the student learn quickly. We assume the teacher cannot directly transfer its knowledge into the student, but instead must use a limited advice budget to tell the student what action to take in a given state. Our work takes a step in this direction where we propose a dual uncertainty based framework to drive advice collection and reuse. On the one hand, the proposed framework allows the student to seek advice from the teacher only when required. On the other hand, the framework reuses teacher's advice wisely by building a model of the teacher and querying it whenever applicable based on the model's confidence.
Much of the existing work in action-advising has focused on either student initiated~\cite{da2020uncertainty}  or teacher initiated ~\cite{da2017simultaneously} criteria to seek advice effectively. Ilhan et al.~\cite{DBLP:conf/atal/IlhanGL21,DBLP:journals/corr/abs-2104-08440} more effectively use an advice budget by having the student build a model of the teacher and use it when possible once the budget has been exhausted. Inspired by these directions, we propose a flexible and systematic uncertainty based framework of effectively seeking and providing advice by using the model of the teacher. Our framework is motivated by real-world teacher-student interaction, where the student directly asks for advice only if he/she is uncertain about a problem. The teacher's prior advice is used by the student to build a mental model of the teacher. Furthermore, when the teacher is unavailable (e.g., the budget is exhausted), the student uses this model of the teacher to solve a problem if it seems familiar or tries to do it on their own otherwise. Our proposed approach is based on such interactions, where the student seeks advice in uncertain regions and the model of the teacher provides advice when its uncertainty is low (states that seem familiar).

This work-in-progress paper has three contributions, introducing and evaluating: (1) An action-advising framework that allows the student to have flexibility in following either its own policy, reusing advice from the model of the teacher, or asking the teacher directly for advice; (2) a methodical uncertainty-based advice reuse approach that leverages the uncertainty of both the student and/or the student's model of the teacher; and (3) a new method of computing the uncertainty for the student agent using a secondary neural network.
To the authors' surprise, while our new methods perform well, they do not outperform state-of-the-art advice reuse methods. Our hope is that this paper will encourage additional work in this area, as we believe there are still many opportunities for significant improvement in how to ask for and reuse, advice from a teacher. 




    
    



\section{Related Work}



 
{\it{Learning from Demonstration}} (LfD) \cite{argall2009survey} and {\it{Action Advising}} \cite{torrey2013teaching} 
are two widely accepted solutions to the sample inefficiency problem in RL algorithms. The sample inefficiency refers to the need of numerous amount of samples to learn a policy for solving a task at hand. This problem is more evident in deep RL where neural networks are used as function approximators and require a large amount of data to converge. LfD is a technique that uses expert demonstrations to bootstrap the learning of agents. LfD is usually an offline process where the expert demonstrations, spanning as long as full trajectories, are provided to the learning agent prior to the start of actual training. Action advising however, is an online technique where an RL agent is provided with actions as advice via another teacher agent or human expert who is optimal or sub-optimal.

\paragraph{Action advising} Action advising paradigm is used extensively in \emph{student-teacher} framework \cite{amir2016interactive,clouse1996integrating,torrey2013teaching,taylor2014reinforcement,da2020uncertainty,DBLP:journals/corr/abs-2104-08440,DBLP:conf/atal/IlhanGL21}. Typically, the student agent is considered a novice that could potentially perform better in the presence of a teacher agent that is pre-trained and known to perform well in the given task. The teacher is available for a limited number of interactions which is usually referred to as the advising budget. Torrey et al.~\cite{torrey2013teaching} introduced action advising along with multiple heuristics for deciding how to provide advice. They mainly introduced \textbf{teacher-driven} advising methods such as {\it Early Advising} and {\it Importance Advising}, among others.\textbf{ Student-driven} advising is also explored in the literature where heuristics like {\it epistemic uncertainty} of the student agent~\cite{da2020uncertainty,Odom2016active} and \textit{advice novelty}~\cite{DBLP:journals/corr/abs-2010-00381} are used to decide when to ask for advice. To reduce the overhead of the teacher, \textbf{ jointly-initiated} action advising was investigated by Amir et al.~\cite{amir2016interactive}, where a student could seek advice  and the teacher could provide advice to the student's queries.  Moreover, the student-teacher framework has also been extended by Da Silva et al.~\cite{da2017simultaneously} to accommodate multiple agents, without having to fix roles of being a student or teacher, and training simultaneously using advice from each other. Lastly, Omidshafiei et al.~\cite{omidshafiei2019learning} used a similar extension of the framework to propose multiple objective functions to learn when and what to advise instead of relying on heuristics.  

\paragraph{Advice reuse} Given the vast literature on action advising, there exists limited work that aims to collect advice for later reuse. This is particularly useful because it allows the student to spend the teacher's budget wisely by not asking redundant or similar queries. 
The idea of reusing collected advice was introduced by Zhu et al.~\cite{zhu2020learning} in a student-teacher framework where advice were reused based on different heuristics such as \textit{QChange}, \textit{Budget reuse}, and \textit{Decay Reusing Probability} in tabular RL algorithms. Due to the tabular nature, they store the action advice in a fixed size buffer for reuse. Our work is heavily inspired by the deep RL compatible work of Ilhan et al.~\cite{DBLP:conf/atal/IlhanGL21,DBLP:journals/corr/abs-2104-08440}, where supervised learning is used to train a model of the teacher from previously collected advice. This model would then predict an action (similar to the teacher) which could be reused if the predetermined probability threshold is met. The limitation of their work includes the student's dependency on the model of the teacher for advice collection. As the student would continue to experience more states in the environment, its measure of uncertainty would account for a larger subset of the state space. We propose a principled framework to address these issues.


\section{Background}

\paragraph{Reinforcement learning} We follow the standard RL  framework~\cite{sutton2018reinforcement} which is modelled as a Markov Decision Process (MDP). It is represented by the tuple $(\mathcal{S},\mathcal{A},R,P,\gamma) $ where $\mathcal{S}$ is the state space, $\mathcal{A}$ is the action space, $R$ is the reward function, $P$ is the state-transition probability and $\gamma \in [0,1)$ is the discount factor for infinite horizon problems. At time-step $t$, an RL agent starts in a state $s_t$, takes an action $a_t$, receives a reward $r_{t+1}$ after interacting with the environment and transitions to the next state $s_{t+1}$. At each time-step, the agent tries maximizing the expected return 
$\mathcal{R}_t=\mathbb{E}[\sum_{k=0}^{T}\gamma^{k}r_{t+k+1}$], which is the discounted sum of rewards that an agent receives when starting from time-step t and following a policy $\pi$ until time step $T$.

\paragraph{Deep Q-Networks} Deep Q-Networks (DQN)~\cite{mnih2015human} is a state-of-the-art off-policy deep RL algorithm to approximate the Q-function for high dimensional tasks with continuous state-space and discrete actions. An underlying neural network is used as a function approximator and the loss function for the iteration $i$ is $L_i(\theta_i)=\mathbb{E}_{s,a \sim P}[y_i-Q(s,a;\theta_i)]^2$ which is the squared error between the target $y_i$ and the Q-value output by the current model parameterized by $\theta_i$. The Bellman equation is used to approximate the target value $y_i=r+\gamma \max_{a'}Q(s',a';\theta_{i-1})$ where $s'$ and $a'$ refers to the next state-action pair, respectively. An experience replay buffer is used to store all the agent interactions. At every training step, a mini-batch of experience is sampled from this buffer to update the parameters of the Q-function. 


\paragraph{Advice reuse using behavior cloning} Imitation learning~\cite{schaal1999imitation} aims at learning the policy of an expert by collecting demonstrations. Behavior Cloning~\cite{pomerleau1991efficient} is a specific approach under  Imitation learning that uses supervised learning to approximate the conditional distributions of actions given the state. Deep neural networks have been used for behavior cloning where the objective is to minimize the negative log-likelihood, $\mathcal{L(\eta)}=\sum_{(s,a) \in \mathcal{T}}-\log M (a|s, \eta)$ where $\mathcal{L(\eta)}$ is the loss-function, $\mathcal{T}$ refers to the demonstration data and $\eta$ is the parameter of the supervised learning model $M$. Ilhan et al. \cite{DBLP:conf/atal/IlhanGL21} used a behavior cloning module to train a neural network using the advice collected from a teacher agent which was referred to as \textit{Advice Imitation}. It was done to enable the student to replicate the teacher's action decisions without needing to query it. Furthermore, epistemic uncertainty via employing dropout regularization in this imitation model lets the student to avoid using the advising budget for similar states for which prior advice had already been taken. We use a similar behavior cloning module for advice reuse in our proposed framework.


\section{Uncertainty-driven Student-Teacher Framework for Advice Collection and Reuse}

This section discusses the problem formulation, introduces the proposed algorithms, and introduces the new way of estimating epistemic uncertainty.

\subsection{Problem formulation}
We propose a methodical student-teacher framework for action advising that aims to wisely reuse advice and minimize redundant advice requests made to the teacher. The focus of this work is to compute and use uncertainty as a metric to decide whether to ask for advice from the teacher, reuse advice from the model of the teacher, or simply let the student continue following its own policy. However, the framework is meant to be general, and any valid metric like \textit{value of information}  \cite{chalkiadakis2003coordination} can be used to drive this decision making 
To formalize the proposed problem:\\
\textbf{Given:} A student agent with its own policy $\pi_S$, a skilled teacher agent $\pi_{T}$, advice budget $b$ which specifies the number of times the teacher $T$ can be queried by the student for action-advice\\
\textbf{To-do:} learn optimal student policy $\pi^*_S$ by leveraging advice from the teacher following $\pi_T$, reusing the model of the teacher build from prior advice $M_{\eta}$ or following its own policy $\pi_S$ subject to the budget $b$

\subsection{Student Uncertainty driven action-advice and reuse }
We propose two algorithms that use uncertainties to drive decision making. In all algorithms, there are two ways to leverage the computed uncertainty. One way is to use a fixed uncertainty threshold to decide whether the student or model of the teacher is uncertain. The other way is to have a dynamically changing threshold based on the previously observed uncertainty values. We keep dynamic uncertainty thresholds for our experiments. The proposed algorithms are as below:

\begin{enumerate}
    \item Student's Uncertainty-driven Advising (SUA)
    \item Student's Uncertainty-driven Advising with Advice Imitation \& Reuse (SUA-AIR)
\end{enumerate}

SUA and SUA-AIR, use the student's uncertainty estimates to drive the advice collection process. Student agent's uncertainty to drive action advising has been explored previously in literature, as mentioned in Section 2. To the best of our knowledge, student uncertainty driven advice collection in an advice-reuse framework hasn't been explored previously. In addition, the process we use to compute the student's uncertainty is novel, which has its advantages, discussed in more detail in Section 4.3.

\begin{figure}[!hbt]
    \centering
    \includegraphics[scale=0.28]{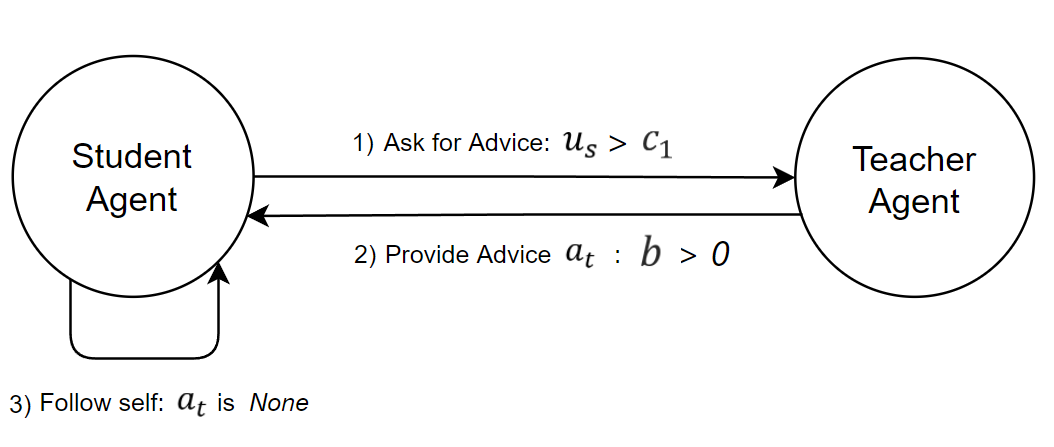}
    \caption{Flow of SUA.}
    \Description{Flow of SUA.}
    \label{fig:SUA}
\end{figure}

The flow of decision-making involved in \textbf{SUA} is shown in Figure \ref{fig:SUA}. At any state, the student agent can ask the teacher agent for advice if its uncertainty $u_{s}$ is greater than the student's adaptive uncertainty threshold $c_1$. In other words, the student can ask the teacher for advice when it is uncertain. The teacher agent can then provide action advice $a_t$ if the teaching budget is not consumed ($b > 0$). Lastly, the student agent would continue to follow its own policy if no advice is received ($a_t$ is None). This is possible if either the student agent is certain in the given state $u_s \leq c_1$, or if the teaching budget is consumed ($b = 0$).

\begin{figure}[!hbt]
    \centering
    \includegraphics[scale=0.28]{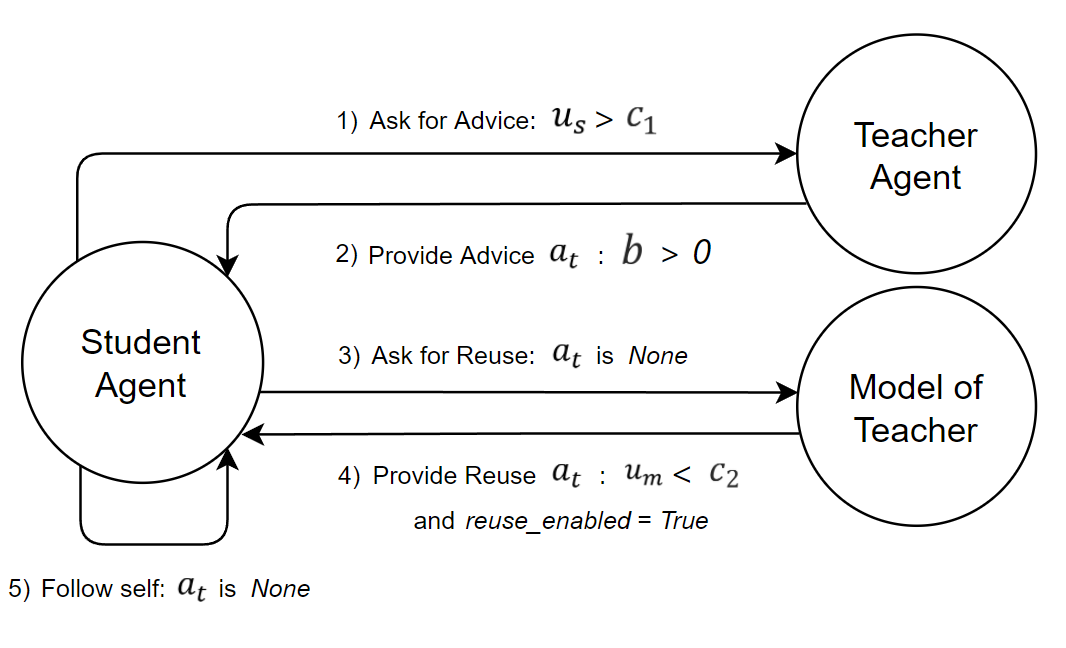}
    \caption{Flow of SUA-AIR.}
    \Description{Flow of SUA-AIR.}
    \label{fig:SUA-AIR}
\end{figure}

\textbf{SUA-AIR}, on the other hand, uses the uncertainty of the student for advice collection and the uncertainty of the model of the teacher for advice reuse. The extent of flexibility available in SUA-AIR and the conditions that trigger those choices are shown in Figure \ref{fig:SUA-AIR}. 
At any state, the student agent can ask the teacher agent for advice if its uncertainty $u_{s}$ is greater than the student's adaptive uncertainty threshold $c_1$. The teacher agent will then provide action advice $a_t$ for the queried state if the teaching budget is not consumed ($b > 0$). The student agent can ask for advice reuse from the model of the teacher if action advice $a_t$ is not received (budget exhausted). The model of the teacher would provide advice if the model uncertainty $u_m$ is less than its uncertainty threshold $c_2$, or simply if the model is certain, and if advice reuse is enabled ($reuse\_enabled = True$). At any step, advice reuse would be enabled if reuse probability $\rho$ is greater than a random probability. Lastly, the student agent would follow its own policy if action $a_t$ is not determined.



The pseudocode for SUA and SUA-AIR is shown in Appendix \ref{sua} and \ref{sua-air}, respectively. SUA serves as a baseline advising method for our experiments. 
SUA-AIR builds upon SUA and AIR to leverage advice reuse with our proposed method of student's uncertainty-driven advising. This allows the student agent to be more independent in asking the teacher for advice, instead of random advice collection or relying on the model of the teacher to get advice as in ~\cite{DBLP:conf/atal/IlhanGL21,DBLP:journals/corr/abs-2104-08440}. Moreover, the uncertainty estimates of the student agent would account for a larger subset of the state-space, enabling more accurate estimates.
\subsection{Computing Student's Epistemic Uncertainty}
The computation of uncertainty occurs at two distinct places, one within the student agent and the other within the model of the teacher. This is evident in Figure \ref{fig:SUA-AIR}, where $u_{s}$ denotes the uncertainty of the student and $u_{m}$ denotes the uncertainty of the model of the teacher.

The way of computing uncertainty is different at each place. As done in \cite{DBLP:journals/corr/abs-2104-08440}, the model of the teacher that is trained in a supervised fashion includes dropout to compute the uncertainty over the action probabilities.
However, using this strategy to compute student agent's uncertainty would come at a cost.
Unless there is a generalisation goal, simply adding a dropout layer in an RL agent can potentially hurt the performance of the learning agent due to the introduction of extra variance that is already plentiful in RL~\cite{DBLP:journals/corr/abs-1810-00123}.
Thus, we propose a new way of computing the RL model's uncertainty by using a secondary neural network with dropout that mimics its learning.

The secondary neural network is trained simultaneously using the samples and targets of the student agent's RL algorithm. 
Since this network is updated over the same data seen by the learning agent, it can be assumed that its uncertainty estimations can serve as a good proxy to the student agent's uncertainty.
Another benefit of having this secondary network approach is that it is easy to decouple.
For example, if the student does not require advising (No Advising) or if the uncertainty estimates are not needed, the secondary network can simply be turned off.

Formally, the uncertainty is computed from the secondary neural network by conducting the following steps. First, $N$ forward passes are performed for a given state $s$. This would give us the matrix $\textbf{F} \in \mathbb{R}^{\:N \times A} $, which contains Q-values for each action $a \in [a_1, a_2, ..., a_A]$, for each forward pass $i \in [1, 2, ..., N]$:
    
    
    \[ \bf{F} = 
    \begin{bmatrix}
    Q_1(s,a_1) \: Q_1(s,a_2) \hdots Q_1(s, a_A) \\
    Q_2(s,a_1) \: Q_2(s,a_2) \hdots Q_2(s, a_A) \\
    \vdots \\
    Q_N(s,a_1) \: Q_N(s,a_2) \hdots Q_N(s,a_A)\\
    \end{bmatrix}\]
    
    where $A$ is the total number of possible actions. The matrix $\bf{F}$ is then used to compute the variance across each column or the Q-values $Q_i(s, a)$ for each action $a$. This results in a row matrix $\textbf{Q} \in \mathbb{R} ^ {:\ 1 \times A}$ :
    
    \[ \bf{Q} = 
    \begin{bmatrix}
    var(Q(s,a_1)) \: var(Q(s,a_2)) \hdots var(Q(s, a_A)) \\
    \end{bmatrix}\]
    
    The average of the values in matrix $\bf{Q}$ then gives us the uncertainty $u_s$:
    
    \[u_s = \frac{\sum_{j=1}^{A} var(Q(s, a_j)) }{A} \]

Computing uncertainty in this manner does not restrict the architecture of the student RL agent to have built in modifications such as dropout or multiple heads. The secondary network, equipped with dropout, is separate from the student RL agent architecture. Moreover, we use adaptive uncertainty thresholds to avoid tuning domain-specific uncertainty thresholds to enable scalability of our methods.

\section{Experiments}
The proposed experiments were designed to answer the following research questions:

\begin{description}
    \item [RQ1] Does adding uncertainties of both the student and the model of the teacher help improve student learning?
    \item [RQ2] How do the proposed algorithms perform against the baselines?
    \item [RQ3] Do the proposed algorithms effect the way teacher advice is requested or reused?
\end{description}


\subsection{Testing Environments}
To evaluate our proposed algorithms against some baseline heuristics in action advising, we have selected five popular domains from the Arcade Learning Environment \cite{bellemare2013arcade}: Pong, Freeway, Seaquest, Enduro, and Q*bert. 

Each environment operates in an RGB pixel space with an observation dimension of $160 \times 120 \times 3$. To reduce the complexities, these observations are pre-processed by converting them to grayscale and shrinking the dimensions to $80 \times 80 \times 1$ via interpolation. Moreover, frames are skipped by repeating the agent actions for 4 consecutive frames to account for the high frame rate. Four resulting frames (or 16 in total) are then stacked upon each other to remove the effect of partial observability, making the final observation dimension to be $80 \times 80 \times 4$.

\subsection{Experimental Setup}
All agents are trained for 5 million training frames and evaluated every 50,000 steps for 10 trials each.
The following agents were tested on each domain for 10 independent runs:


\begin{description}
    \item [No Advising (NA):] A student agent with no advising (no teacher).   
    \item [Early Advising (EA):] A student agent with advising provided in early phase of training until the the teaching budget exhausts. 
    \item [Random Advising (RA):] A student agent advised with $0.5$ probability at every step until the budget runs out.
    \item [Advice Imitation \& Reuse (AIR):] Previous state-of-the-art baseline \cite{DBLP:journals/corr/abs-2104-08440} that uses the uncertainty of the model of the teacher to drive advice collection and reuse. 
    \item [Student's Uncertainty-driven Advising (SUA):] A student agent that uses its adaptive uncertainty estimates to drive advice collection; no advice reuse. 
    \item [Student's Uncertainty-driven Advising with] \textbf{Advice Imitation \& Reuse (SUA-AIR):} A student agent that uses its adaptive uncertainty estimates to drive advice collection; paired with a teacher imitation model that uses its adaptive uncertainty thresholds for advice reuse.

\end{description}


\begin{figure*}[!hbt]
    \centering
    \includegraphics[scale=0.37]{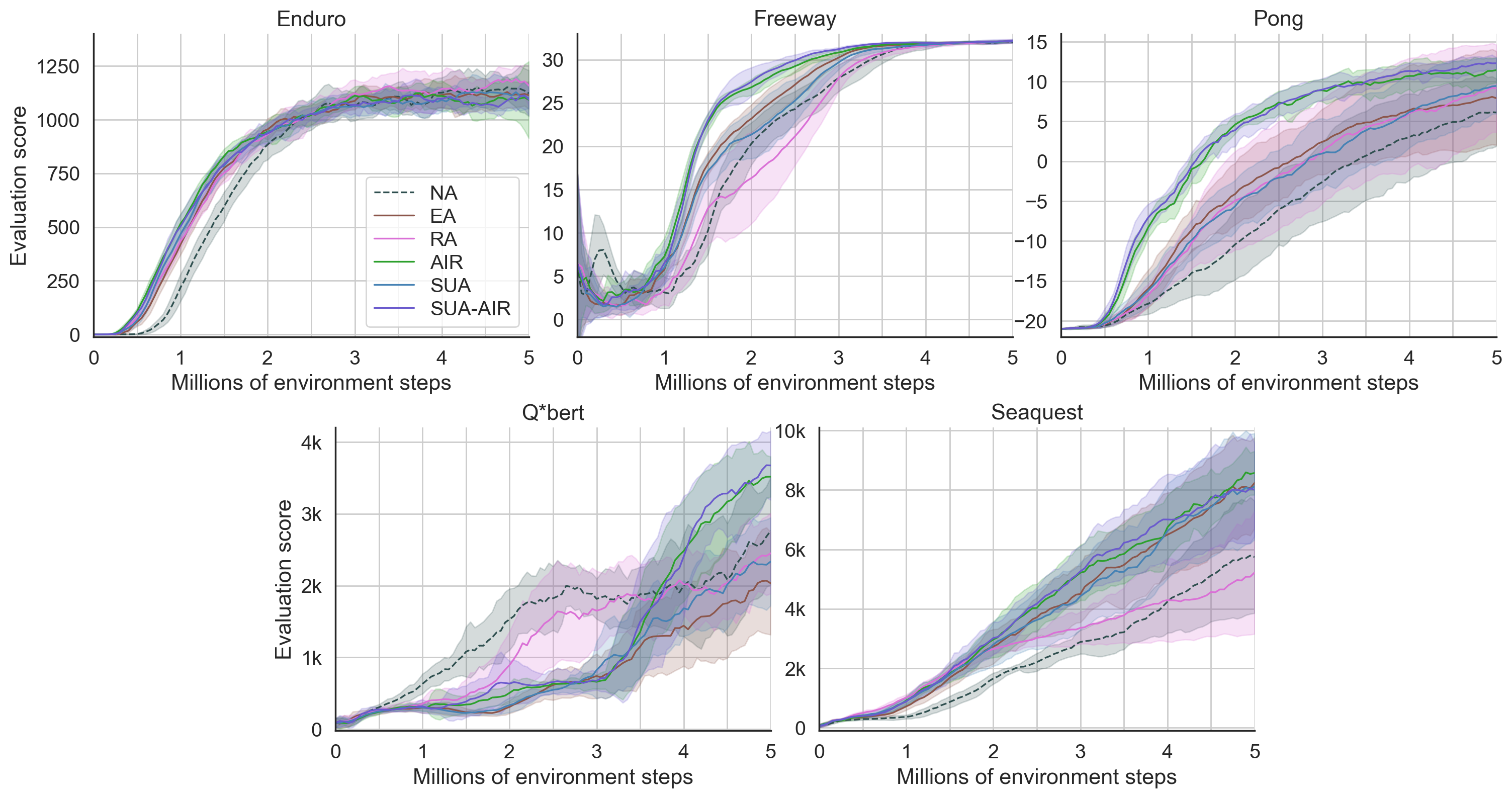}
    \caption{Evaluation rewards for all agents tested on Enduro, Freeway, Pong, Q*bert, and Seaquest.}
    \Description{Evaluation curves showing the rewards for all agents tested on Enduro, Freeway, Pong, Q*bert, and Seaquest.}
    \label{fig:eval_scores}
\end{figure*}

\begin{figure*}[!hbt]
    \centering
    \includegraphics[scale=0.38]{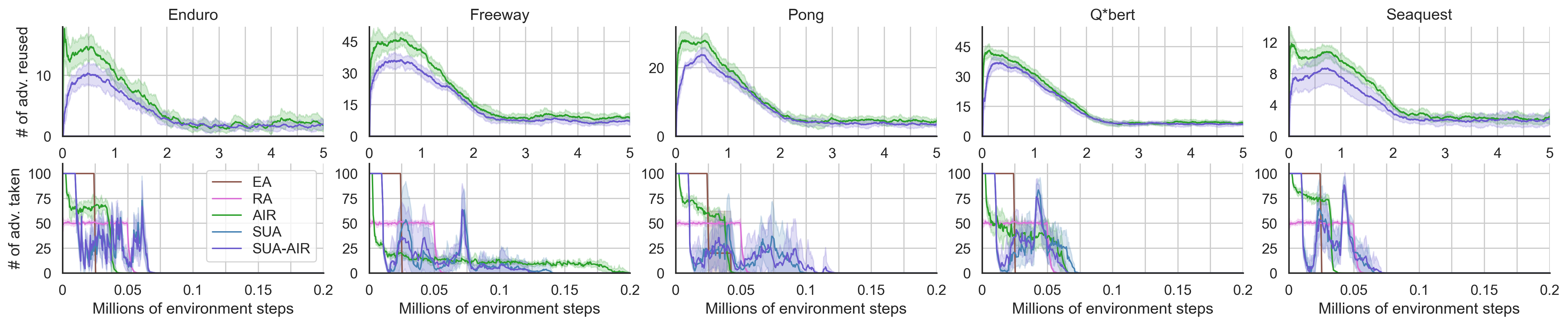}
    \caption{Advice reuse (actions taken from the model of the teacher) and advice taken (actions taken directly from the teacher agent) in every 100 steps taken by the student in all domains.}
    \Description{A combined plot showing advice reuse (actions taken from the model of the teacher) and advice taken (actions taken directly from the teacher agent) in every 100 steps taken by the student agents in all domains.}
    \label{fig:advice_schedule}
\end{figure*}

All student agents use the same architecture, Double DQN with 3 convolutional layers followed by a fully-connected hidden layer and a dueling output. All agents use the $\epsilon$-greedy strategy for exploration where $\epsilon$ is decayed over time. The secondary network's structure is similar to student DQN with two added differences. First, the secondary network is a supervised learning model, and second, has a dropout layer.\footnote{Following the literature \cite{chen2017agent}, the dropout rate is set to 0.2 so that it does not hamper the learning progress.} Dropout rate controls the percentage of units that are dropped at every training step and the number of forward passes performed to compute the epistemic uncertainty is set to 100.

The supervised learning model of teacher (or imitation model), trained with the student-teacher interaction data, is equipped with the network structure identical to the student agent's secondary network. Here, the model of teacher predicts action probabilities (instead of Q-values in the secondary network). A dropout layer is also added to the model. Following AIR \cite{DBLP:journals/corr/abs-2104-08440}, the dropout rate is 0.35. Again, the number of forward passes to compute the epistemic uncertainty is set to 100. The list of all the hyperparameters for the student agents is shown in Appendix \ref{appendix:hyperparam-students} and the list of all the hyperparameters for the model of the teacher (or imitation model) is shown in Appendix \ref{appendix:hyperparam-model}.

For each game, a teacher agent is pre-trained and has a fixed policy. 
The teacher agents have the same network structure and algorithm as the student agent.
The teacher agents can be considered competent, as compared to results in the literature, and obtain average evaluation scores of 1556 for Enduro, 28.8 for Freeway, 12 for Pong, 3705 for Q*bert, and 8178 for Seaquest. 

\section{Results}
The evaluation performance of all the agents in Pong, Freeway, Seaquest, Enduro, and Q*bert is reported in Figure \ref{fig:eval_scores}. All advised agents (except RA) performed better than no advising agent (NA) during the early training phases, with statistically significant differences as they are more than twice the standard error in the respective means (for 95\% confidence), in all domains except Q*bert,  where the benefits of advising appear in later stages of training.


In Enduro, the performance of all advised agents during the early phases of training was better than NA agent with statistically significant differences. The benefits of advice reuse become slightly more apparent when we look at the performance of agents in Freeway. Students with advice reuse, SUA-AIR and AIR, showed a statistically significant boost in performance during early-to-mid training, as compared to other agents. Advising methods such as EA, RA, and SUA took more time to catch up to the policies of students using advice reuse in Freeway. The difference of performance between students with and without advice reuse becomes more evident in Pong. Students with advice reuse (SUA-AIR and AIR) showed a statistically significant boost in performance throughout the majority of training (other than the end of training). Advised agents without advice reuse (EA, SUA, and RA) failed to keep up with the performance of advice reuse students in Pong. In Q*bert, the advice reuse agents (SUA-AIR and AIR) performed statistically better than all other agents towards the end of training. In Seaquest, most of the advised agents (except RA) took the lead over the NA agent throughout the majority of training (other than the end of training) with statistically significant differences.

In general, SUA-AIR and AIR performed in a similar fashion. The differences in performance between SUA-AIR and AIR are not statistically significant. Thus, the answer to RQ1 is inconclusive. However, SUA-AIR does perform statistically better in various phases of training (e.g. early, later phases) across different domains (e.g. Pong, Q*bert), as compared to other agents (except AIR).
A table reporting the evaluation scores of the student agents across all domains is shown in Appendix \ref{table:Eval_Scores}.

The advice taken and the reuse schedule are shown in Figure \ref{fig:advice_schedule}. The number of advice reused (top row) for AIR is higher than SUA and SUA-AIR in almost all steps. However, having a higher advice reuse rate does not necessarily correspond to a better evaluation performance. The amount of advice taken (bottom row) for SUA and SUA-AIR is, in general, spread in an erratic fashion across the environment steps. AIR seeks to consume most of the training budget early on due to the higher uncertainty of the model of the teacher. Whereas the proposed algorithms do a better job of asking the teacher as and when needed. Thus, to answer RQ3, the proposed algorithms do bring a change in the way advice is taken directly from the teacher.

\subsection*{Evaluating Model Performance} 
\begin{figure}[!hbt]
        \centering
        \includegraphics[scale=0.45]{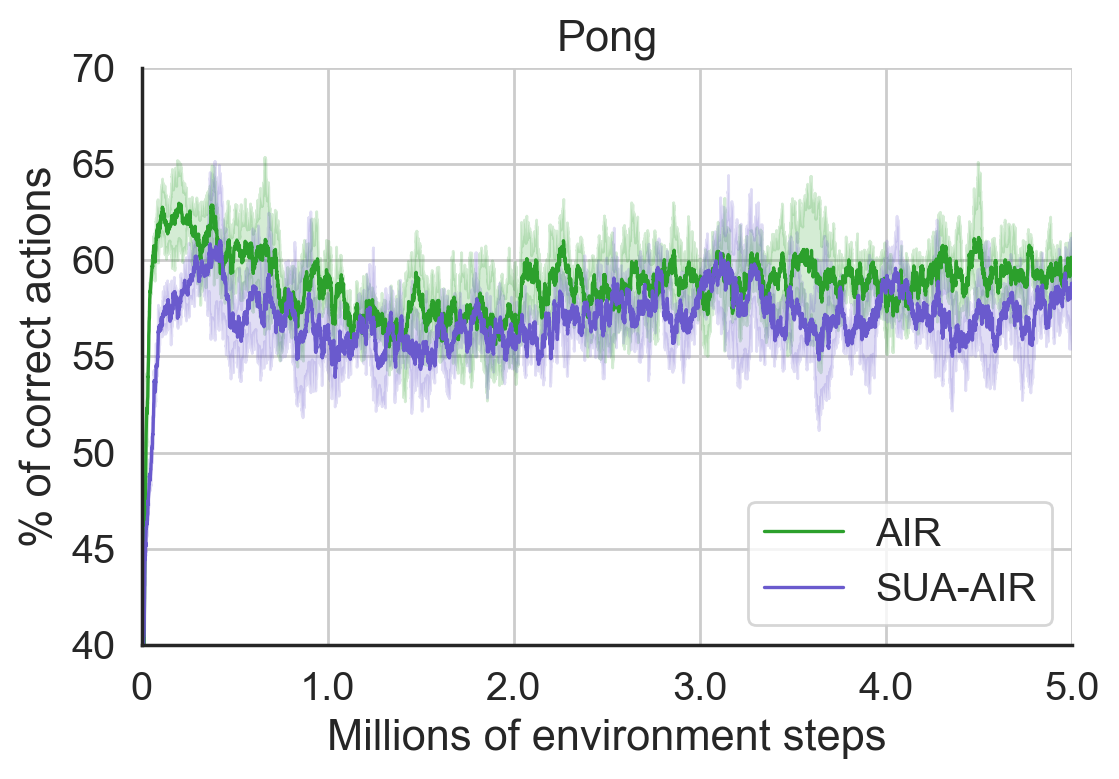}
        \caption{Percentage of correct actions taken by the model of the teacher for SUA-AIR and AIR across training steps in Pong averaged over 3 independent runs.}
        \Description{A plot showing the percentage of correct actions taken by the model of the teacher for SUA-AIR and AIR across training steps in Pong averaged over 3 independent runs.}
        \label{fig:Model_Accuracy}
    \end{figure}

To further investigate the similar evaluation performance of SUA-AIR and AIR, we evaluate the accuracy of the model of teacher by comparing the actions of the teacher and the model for the states that the student visits. This evaluation is shown in Figure \ref{fig:Model_Accuracy}. 

\noindent The percentage of correct actions taken by the model (y-axis), correct actions are actions that are the same as teacher's, over the course of millions of environment steps (x-axis) taken by the student agent in Pong are reported. It is evident from Figure \ref{fig:Model_Accuracy} that the model of the teacher for SUA-AIR and AIR show similar accuracies across training. Since AIR uses the model of the teacher to collect advice, it is natural to see it slightly more accurate than the model of SUA-AIR. However, this does not impact the evaluation performance of SUA-AIR. We expected SUA-AIR to be better than AIR, but it is not, and we intend to understand why in the future work. Due to the similar performance of these models of the teacher, SUA and SUA-AIR showed similar evaluation performance across all games.
 
\section{Conclusion and Future Work}
We proposed two new methods, Student's Uncertainty-driven Advising (SUA) and Student's Uncertainty-driven Advising with Advice Imitation \& Reuse (SUA-AIR). Both SUA and SUA-AIR use our proposed method for computing student's uncertainty to drive the advice collection process. This uncertainty of the student agent and the uncertainty of the model of teacher were then used for the advice collection (in SUA and SUA-AIR) and advice reuse processes (in SUA-AIR), respectively. The model of teacher was trained using advice interaction data between the student and teacher agents. Using this framework, the student agent can decide when to ask the teacher agent for direct advice, or the model of teacher for advice reuse, or when to follow the student's own policy. We hoped that SUA-AIR would outperform AIR, but we found that the differences were not statistically significant. Moreover, the results show that using advice reuse, in action advising RL agents, provides a significant boost in performance in different stages of training. 

There are multiple avenues for future work. Currently, the student agent in SUA-AIR considerably leverages the model of the teacher for advice reuse after the teaching budget is consumed. This can be further extended to add more flexibility where the student agent could start asking the model of the teacher for reuse during the consumption of the teaching budget to use the teaching budget efficiently. Moreover, the decision to reuse advice in SUA-AIR precedes the decision of the student agent. This can be extended to account for the student agent's uncertainty before reusing advice from the model of the teacher. Furthermore, a more thorough study could be conducted to test different advice reuse schedules by modifying the initial reuse probability, final reuse probability, and the total decaying steps. For example, the final reuse probability could be set to 0 to allow the student agent to become independent from the model of the teacher towards the later stages of training. Lastly, we currently use fixed percentile values to compute the uncertainties for the student agent and the model of teacher. This could be better extended to follow a dynamic schedule where the percentile values start closer to 50 in early training and then restricted to values closer to 100 in the later stages of training. This change would be better suited for advice reuse since a lower percentile value for the student agent's uncertainty would not capture the states for which the student is genuinely uncertain.

\section*{Acknowledgements}
This work has taken place in the Intelligent Robot Learning Lab at the University of Alberta, which is supported in part by research grants from the AI4Society; the Alberta Machine Intelligence Institute (Amii); a Canada CIFAR AI Chair, Amii; Compute Canada; and NSERC.

\balance
\newpage
\bibliographystyle{ACM-Reference-Format} 
\bibliography{sample}

\newpage
\section*{Appendix}
    

Since SUA, SUA-AIR, and AIR use adaptive uncertainty thresholds, a hyperparameter for fixing the uncertainty threshold is not required. However, all of the mentioned algorithms require one or two percentile values to determine the uncertainty thresholds automatically. The percentile $p_1$ for the student's uncertainty in SUA and SUA-AIR is set to 70 to maintain a balance for the student agent in asking the teacher for advice. Whereas percentile $p_2$ for the uncertainty of the model of teacher in SUA-AIR and AIR is set to 90. For SUA-AIR and AIR, the initial probability of reuse $\rho_{init}$ is set to 0.5, and the final probability of reuse $\rho_{final}$ is set to 0.1. The decay from $\rho_{init}$ to $\rho_{final}$ happens in a total of 1.5 million steps, starting from 500,00 steps and ending at 2.5 million steps. For training the model of teacher or the imitation model (in SUA-AIR and AIR), the minimum steps $t_{min}$ is set to 50,000, and the minimum samples $n_{min}$ is set to 2,500. Moreover, the initial training iterations $k_{init}$ and periodic training iterations $k_{periodic}$ for the model of teacher (in SUA-AIR and AIR) are set to 50,000 and 20,000, respectively. Lastly, the {\it min-window-size} for the uncertainty buffer $D_u$ is set to 200, and the maximum size is set to 10,000. The values for minimum and maximum window size for the uncertainty buffer were selected based on general performance across all domains. The values for all hyperparameters pertaining to the student agent and the imitation model are listed in Table \ref{appendix:hyperparam-students} and \ref{appendix:hyperparam-model}. The values for all the mentioned hyperparameters are kept the same across experiments in all domains. 
    

All other hyperparameter values except replay memory minimum size and $\epsilon$ decaying steps are kept the same as in previous work \cite{DBLP:journals/corr/abs-2104-08440}. Minimum replay memory size is decreased to 10k (from 50k) to enable early training of the student agent (from the replay buffer) to produce accurate uncertainty estimations and $\epsilon$ decaying steps are decreased to 250k (from 500k) to limit exploration in later stages of training.

We also note that $min\_window\_size$ for $D_u$ (in Table~\ref{appendix:hyperparam-students}) is set to 200 to initiate the computation of the dynamic threshold $c_1$ and to ensure that there is enough data to compute accurate estimates. This will have very little impact as we will reach this value within the first episode.

\begin{table*}[pht!]
    \begin{tabular}{ll|lllll|}
    \cline{3-7}
    {\ul }                         &         & \multicolumn{5}{c|}{Evaluation Scores}                                                                                                     \\ \hline
    \multicolumn{1}{|l|}{Domain}   & Student & \multicolumn{1}{l|}{Initial (1/3)}                            & \multicolumn{1}{l|}{Inter. (2/3)}                         & \multicolumn{1}{l|}{Later (3/3)}                                 & \multicolumn{1}{l|}{Final}                                      & Total                                        \\ \hline
    \multicolumn{1}{|l|}{}         & NA      & \multicolumn{1}{l|}{196.63 $\pm$ 9.47}           & \multicolumn{1}{l|}{975.27 $\pm$ 18.70}            & \multicolumn{1}{l|}{1120.26 $\pm$ 17.31}           & \multicolumn{1}{l|}{1133.44 $\pm$ 33.07}           & 769.67 $\pm$ 12.02              \\
    \multicolumn{1}{|l|}{}         & EA      & \multicolumn{1}{l|}{316.59 $\pm$ 9.29}           & \multicolumn{1}{l|}{1011.35 $\pm$ 8.15}            & \multicolumn{1}{l|}{1113.29 $\pm$ 15.44}           & \multicolumn{1}{l|}{1116.52 $\pm$ 17.57}           & 818.67 $\pm$ 6.99               \\
    \multicolumn{1}{|l|}{Enduro}   & RA      & \multicolumn{1}{l|}{320.86 $\pm$ 5.56}           & \multicolumn{1}{l|}{\textbf{1015.65 $\pm$ 13.25}}  & \multicolumn{1}{l|}{\textbf{1148.66 $\pm$ 17.64}}  & \multicolumn{1}{l|}{\textbf{1169.99 $\pm$ 25.13}}  & \textbf{833.41 $\pm$ 11.05}     \\
    \multicolumn{1}{|l|}{}         & AIR     & \multicolumn{1}{l|}{\textbf{374.08 $\pm$ 5.07}}  & \multicolumn{1}{l|}{1014.47 $\pm$ 7.82}            & \multicolumn{1}{l|}{1094.33 $\pm$ 16.12}           & \multicolumn{1}{l|}{1097.87 $\pm$ 47.19}           & 832.12 $\pm$ 8.11               \\
    \multicolumn{1}{|l|}{}         & SUA     & \multicolumn{1}{l|}{341.69 $\pm$ 6.71}           & \multicolumn{1}{l|}{1000.59 $\pm$ 11.22}           & \multicolumn{1}{l|}{1102.97 $\pm$ 10.42}           & \multicolumn{1}{l|}{1106.68 $\pm$ 17.35}           & 819.77 $\pm$ 7.66               \\
    \multicolumn{1}{|l|}{}         & SUA-AIR & \multicolumn{1}{l|}{359.34 $\pm$ 7.33}           & \multicolumn{1}{l|}{1003.02 $\pm$ 6.80}            & \multicolumn{1}{l|}{1089.54 $\pm$ 11.23}           & \multicolumn{1}{l|}{1105.11 $\pm$ 19.09}           & 821.84 $\pm$ 6.01               \\ \hline
    \multicolumn{1}{|l|}{}         & NA      & \multicolumn{1}{l|}{5.51 $\pm$ 0.28}             & \multicolumn{1}{l|}{23.68 $\pm$ 0.61}              & \multicolumn{1}{l|}{31.57 $\pm$ 0.13}              & \multicolumn{1}{l|}{32.04 $\pm$ 0.04}              & 20.40 $\pm$ 0.28                \\
    \multicolumn{1}{|l|}{}         & EA      & \multicolumn{1}{l|}{6.82 $\pm$ 0.40}             & \multicolumn{1}{l|}{26.58 $\pm$ 0.35}              & \multicolumn{1}{l|}{31.87 $\pm$ 0.03}              & \multicolumn{1}{l|}{32.11 $\pm$ 0.04}              & 21.91 $\pm$ 0.22                \\
    \multicolumn{1}{|l|}{Freeway}  & RA      & \multicolumn{1}{l|}{5.01 $\pm$ 0.59}             & \multicolumn{1}{l|}{21.55 $\pm$ 1.03}              & \multicolumn{1}{l|}{31.58 $\pm$ 0.07}              & \multicolumn{1}{l|}{32.09 $\pm$ 0.07}              & 19.52 $\pm$ 0.53                \\
    \multicolumn{1}{|l|}{}         & AIR     & \multicolumn{1}{l|}{\textbf{8.95 $\pm$ 0.38}}    & \multicolumn{1}{l|}{28.84 $\pm$ 0.16}              & \multicolumn{1}{l|}{31.89 $\pm$ 0.03}              & \multicolumn{1}{l|}{32.12 $\pm$ 0.05}              & 23.37 $\pm$ 0.13                \\
    \multicolumn{1}{|l|}{}         & SUA     & \multicolumn{1}{l|}{6.72 $\pm$ 0.24}             & \multicolumn{1}{l|}{25.29 $\pm$ 0.60}              & \multicolumn{1}{l|}{31.75 $\pm$ 0.05}              & \multicolumn{1}{l|}{32.10 $\pm$ 0.04}              & 21.40 $\pm$ 0.25                \\
    \multicolumn{1}{|l|}{}         & SUA-AIR & \multicolumn{1}{l|}{8.57 $\pm$ 0.43}             & \multicolumn{1}{l|}{\textbf{29.36 $\pm$ 0.23}}     & \multicolumn{1}{l|}{\textbf{31.98 $\pm$ 0.02}}     & \multicolumn{1}{l|}{\textbf{32.15 $\pm$ 0.07}}     & \textbf{23.45 $\pm$ 0.18}       \\ \hline
    \multicolumn{1}{|l|}{}         & NA      & \multicolumn{1}{l|}{-18.37 $\pm$ 0.29}           & \multicolumn{1}{l|}{-6.70 $\pm$ 1.36}              & \multicolumn{1}{l|}{3.51 $\pm$ 1.59}               & \multicolumn{1}{l|}{6.12 $\pm$ 1.39}               & -7.08 $\pm$ 1.01                \\
    \multicolumn{1}{|l|}{}         & EA      & \multicolumn{1}{l|}{-16.58 $\pm$ 0.37}           & \multicolumn{1}{l|}{-1.02 $\pm$ 1.09}              & \multicolumn{1}{l|}{6.43 $\pm$ 1.73}               & \multicolumn{1}{l|}{7.97 $\pm$ 1.87}               & -3.59 $\pm$ 0.91                \\
    \multicolumn{1}{|l|}{Pong}     & RA      & \multicolumn{1}{l|}{-17.17 $\pm$ 0.36}           & \multicolumn{1}{l|}{-2.08 $\pm$ 1.92}              & \multicolumn{1}{l|}{6.40 $\pm$ 2.08}               & \multicolumn{1}{l|}{9.11 $\pm$ 1.76}               & -4.16 $\pm$ 1.39                \\
    \multicolumn{1}{|l|}{}         & AIR     & \multicolumn{1}{l|}{-12.52 $\pm$ 0.24}           & \multicolumn{1}{l|}{\textbf{6.46 $\pm$ 0.30}}      & \multicolumn{1}{l|}{10.62 $\pm$ 0.45}              & \multicolumn{1}{l|}{11.36 $\pm$ 0.51}              & 1.66 $\pm$ 0.23                 \\
    \multicolumn{1}{|l|}{}         & SUA     & \multicolumn{1}{l|}{-16.89 $\pm$ 0.34}           & \multicolumn{1}{l|}{-2.50 $\pm$ 1.13}              & \multicolumn{1}{l|}{6.60 $\pm$ 1.30}               & \multicolumn{1}{l|}{9.35 $\pm$ 1.12}               & -4.14 $\pm$ 0.86                \\
    \multicolumn{1}{|l|}{}         & SUA-AIR & \multicolumn{1}{l|}{\textbf{-11.80 $\pm$ 0.28}}  & \multicolumn{1}{l|}{6.39 $\pm$ 0.38}               & \multicolumn{1}{l|}{\textbf{11.19 $\pm$ 0.20}}     & \multicolumn{1}{l|}{\textbf{12.32 $\pm$ 0.27}}     & \textbf{2.06 $\pm$ 0.19}        \\ \hline
    \multicolumn{1}{|l|}{}         & NA      & \multicolumn{1}{l|}{\textbf{523.17 $\pm$ 22.72}}          & \multicolumn{1}{l|}{\textbf{1709.45 $\pm$ 39.80}}           & \multicolumn{1}{l|}{2127.62 $\pm$ 182.23}          & \multicolumn{1}{l|}{2678.62 $\pm$ 225.15}          & \textbf{1462.62 $\pm$ 67.15}             \\
    \multicolumn{1}{|l|}{}         & EA      & \multicolumn{1}{l|}{239.70 $\pm$ 12.52}          & \multicolumn{1}{l|}{539.89 $\pm$ 49.23}            & \multicolumn{1}{l|}{1544.32 $\pm$ 157.99}          & \multicolumn{1}{l|}{2054.10 $\pm$ 225.66}          & 779.93 $\pm$ 64.51              \\
    \multicolumn{1}{|l|}{Q*bert}   & RA      & \multicolumn{1}{l|}{308.00 $\pm$ 23.54} & \multicolumn{1}{l|}{1332.33 $\pm$ 150.13} & \multicolumn{1}{l|}{2027.82 $\pm$ 115.96}          & \multicolumn{1}{l|}{2427.65 $\pm$ 165.35}          & 1231.77 $\pm$ 78.08    \\
    \multicolumn{1}{|l|}{}         & AIR     & \multicolumn{1}{l|}{263.38 $\pm$ 20.06}          & \multicolumn{1}{l|}{610.25 $\pm$ 25.74}            & \multicolumn{1}{l|}{2588.68 $\pm$ 148.04}          & \multicolumn{1}{l|}{3508.49 $\pm$ 97.30}           & 1162.92 $\pm$ 56.82             \\
    \multicolumn{1}{|l|}{}         & SUA     & \multicolumn{1}{l|}{233.19 $\pm$ 7.53}           & \multicolumn{1}{l|}{571.69 $\pm$ 51.65}            & \multicolumn{1}{l|}{1797.56 $\pm$ 116.56}          & \multicolumn{1}{l|}{2307.20 $\pm$ 191.48}          & 873.76 $\pm$ 54.32              \\
    \multicolumn{1}{|l|}{}         & SUA-AIR & \multicolumn{1}{l|}{270.13 $\pm$ 17.40}          & \multicolumn{1}{l|}{660.25 $\pm$ 60.88}            & \multicolumn{1}{l|}{\textbf{2641.93 $\pm$ 185.13}} & \multicolumn{1}{l|}{\textbf{3653.54 $\pm$ 151.78}} & 1199.88 $\pm$ 78.84             \\ \hline
    \multicolumn{1}{|l|}{}         & NA      & \multicolumn{1}{l|}{407.18 $\pm$ 21.80}          & \multicolumn{1}{l|}{2177.36 $\pm$ 40.02}           & \multicolumn{1}{l|}{4503.52 $\pm$ 378.01}          & \multicolumn{1}{l|}{5747.69 $\pm$ 606.93}          & 2382.05 $\pm$ 127.50          \\
    \multicolumn{1}{|l|}{}         & EA      & \multicolumn{1}{l|}{703.16 $\pm$ 27.12}          & \multicolumn{1}{l|}{3628.69 $\pm$ 200.65}          & \multicolumn{1}{l|}{6799.10 $\pm$ 452.63}          & \multicolumn{1}{l|}{8131.73 $\pm$ 487.51}          & 3740.09 $\pm$ 205.78         \\
    \multicolumn{1}{|l|}{Seaquest} & RA      & \multicolumn{1}{l|}{\textbf{900.46 $\pm$ 11.64}} & \multicolumn{1}{l|}{2973.80 $\pm$ 94.90}           & \multicolumn{1}{l|}{4357.73 $\pm$ 426.44}          & \multicolumn{1}{l|}{5126.56 $\pm$ 628.49}          & 2762.25 $\pm$ 165.89          \\
    \multicolumn{1}{|l|}{}         & AIR     & \multicolumn{1}{l|}{814.67 $\pm$ 27.28}          & \multicolumn{1}{l|}{4009.80 $\pm$ 202.07}          & \multicolumn{1}{l|}{7099.48 $\pm$ 258.61}          & \multicolumn{1}{l|}{\textbf{8569.04 $\pm$ 231.59}} & 4005.94 $\pm$ 146.88          \\
    \multicolumn{1}{|l|}{}         & SUA     & \multicolumn{1}{l|}{806.88 $\pm$ 19.11}          & \multicolumn{1}{l|}{3571.92 $\pm$ 266.72}          & \multicolumn{1}{l|}{6751.58 $\pm$ 483.42}          & \multicolumn{1}{l|}{8080.32 $\pm$ 581.71}          & 3738.87 $\pm$ 240.52          \\
    \multicolumn{1}{|l|}{}         & SUA-AIR & \multicolumn{1}{l|}{819.10 $\pm$ 31.64}          & \multicolumn{1}{l|}{\textbf{4091.41 $\pm$ 179.81}} & \multicolumn{1}{l|}{\textbf{7135.82 $\pm$ 449.44}} & \multicolumn{1}{l|}{8000.41 $\pm$ 541.18}          & \textbf{4047.09 $\pm$ 206.70} \\ \hline
    \end{tabular}
    \caption{Evaluation scores with respect to initial, intermediate, later, final, and total performance of all agents in 5 domains averaged over 10 independent runs. The standard errors are reported with $\pm$. The best scores are reported in bold.}
    \label{table:Eval_Scores}
\end{table*}

    \begin{algorithm*}[pht!]
         \footnotesize \caption{\underline{\textbf{S}}tudent's \underline{\textbf{U}}ncertainty-driven \underline{\textbf{A}}dvising}
        \label{sua}
        \begin{algorithmic}[1]
        \State \textbf{Input}: max training iterations $t_{max}$, teacher agent policy $\pi_{E}$, student DQN policy $\pi_{S}$, 
        max advising budget $b$, secondary network $H_{\omega}$, minimum window size for uncertainty {\it min-window-size}
        \State $D_{dqn}\gets \emptyset$; $D_{u}\gets \emptyset$ \Comment{initialize replay and uncertainty buffers, respectively}
        \State $c_1 \gets None$; \Comment{initialize uncertainty threshold}
        \For {training steps $t \in \{1,2,\cdots, t_{max}\}$}
        \If {$Env$ is reset}
          \State Obtain start state $s_t$ from $Env$
        \EndIf
        \State $a_t \gets None$ \Comment{action not determined yet}
        \LeftComment{\textbf{\textcolor{black}{Advice Collection}}}
        \If {DQN-student-model is trained at least once}
        \State $u_s \gets H_{\omega}^u(s_t)$ \Comment{measure student uncertainty}
        \State $D_u$.append($u_s$)
        \EndIf
        \If {$b>0$}
        \If {DQN-student-model is trained at least once}
        \If {$|D_u| \geq$ {\it min-window-size}}
        \State $c_1 \gets$ determine threshold using $D_u$
        \Else
        \State $c_1 \gets -\infty$ \Comment{activate early advising}
        \EndIf
        \If {$u_s > c_1$} \Comment{uncertainty-based advising or early advising}
        \State $a_t \gets \pi_{E}(s_t)$ 
        \State $b \gets b-1$
        \EndIf
        \Else
        \State $a_t \gets \pi_{E}(s_t)$ \Comment{early advising}
        \State $b \gets b-1$
        \EndIf
        \EndIf
        \LeftComment{\textbf{\textcolor{black}{Self Policy (Ilhan et al., 2021)}}}
        \If {$a_t$ is None}
        \State $a_t \gets \pi_{S}(s_t)$ \Comment{follow student's current policy}
        \EndIf
        \State Take action $a_t$, obtain $s_{t+1},r_{t}$ from $Env$
        \State $D_{dqn}$.append($\left \langle s_t,a_t,r_t,s_{t+1}\right \rangle$) \Comment {Store experience}
        \State Update DQN-student-model
        \State Update $H_{\omega}$ with DQN's mini-batch \& target
        \State $s_t \gets s_{t+1}$
        \EndFor
        \end{algorithmic}
    \end{algorithm*}
  
    \begin{algorithm*}[ht!]
        \footnotesize \caption{Advice Imitation Model}
        \label{advice-imitation}
        \begin{algorithmic}[1]
        \Function{BuildAdvModel}{$D,M_{\eta},n_{last},n_{min},t_{min},t,t_{last},k_{init},k_{periodic}$}
        \If {($|D|-n_{last} >=n_{min}$ \textbf{or}\\
           $~~~~~~~~~(|D|-n_{last} >=n_{min}/2$ \textbf{and} $t-t_{last} >=t_{min})$} \Comment{check if the model can be trained}
            \State Train $M_\eta$ using $D$ for $k_{init}$ or $k_{periodic}$ iterations
            \Comment{train the imitation model}
            \State Determine $c_2$ as done in AIR  \cite{DBLP:journals/corr/abs-2104-08440}
            \Comment{compute the uncertainty threshold}
            \State $n_{last} \gets |D|$
            \State $t_{last} \gets t$
        \EndIf
        
        \State \Return $M_{\eta}, c_2, n_{last}, t_{last}$
        \EndFunction
        \end{algorithmic}
    \end{algorithm*}

    
    \begin{algorithm*}[ht!]
        \footnotesize \caption{\underline{\textbf{S}}tudent's \underline{\textbf{U}}ncertainty-driven \underline{\textbf{A}}dvising with \underline{\textbf{A}}dvise \underline{\textbf{I}}mitation and \underline{\textbf{R}}euse}
        \label{sua-air}
        \begin{algorithmic}[1]
        \State \textbf{Input}: max. training iterations $t_{max}$, teacher agent policy $\pi_{E}$, student DQN policy $\pi_{S}$, advice model $M_{\eta}$, advising budget $b$, secondary network $H_{\omega}$, minimum window size for uncertainty {\it min-window-size}, initial reuse probability $\rho_{init}$, final reuse probability $\rho_{final}$, total $\rho$  decaying steps $t_{\rho}$, 
        min. samples collected and steps taken to begin training $M_\eta$: $n_{min}$, $t_{min}$ respectively, initial imitation training iterations $k_{init}$, periodic imitation training iterations $k_{periodic}$
        \State $D \gets \emptyset$; $D_{dqn}\gets \emptyset$; $D_{u}\gets \emptyset$ \Comment{initialize advice, replay, and uncertainty buffers, respectively}
        \State $c_1 \gets None$; $c_2 \gets None$ \Comment{initialize student and model uncertainty thresholds, respectively}
        \State $n_{last} \gets 0$; $t_{last} \gets 0$
        \State $\rho \leftarrow \rho_{init}$ \Comment{set reuse probability with initial value}
        \For {training steps $t \in \{1,2,\cdots, t_{max}\}$}
        \If {$Env$ is reset}
          \State $reuse\_enabled \gets True$ with probability $\rho$, $False$ otherwise
          \State Obtain start state $s_t$ from $Env$
        \EndIf
        \State $a_t \gets None$ \Comment{action not determined yet}
        \LeftComment{\textbf{\textcolor{black}{Advice Collection}}}
        \If {DQN-student-model is trained at least once}
        \State $u_s \gets H_{\omega}^u(s_t)$ \Comment{measure student uncertainty}
        \State $D_u$.append($u_s$)
        \EndIf
        \If {$b>0$}
        \If {DQN-student-model is trained at least once}
        \If {$|D_u| \geq$ {\it min-window-size}}
        \State $c_1 \gets$ determine threshold using $D_u$
        \Else
        \State $c_1 \gets -\infty$ \Comment{activate early advising}
        \EndIf
        \If {$u_s > c_1$} \Comment{uncertainty-based advising or early advising}
        \State $a_t \gets \pi_{E}(s_t)$
        \State $D.$append($\left \langle s_t,a_t \right \rangle$) \Comment{append to advice buffer}
        \State $b \gets b-1$
        \EndIf
        \Else
        \State $a_t \gets \pi_{E}(s_t)$ \Comment{early advising}
        \State $D.$append($\left \langle s_t,a_t \right \rangle$) \Comment{append to advice buffer}
        \State $b \gets b-1$
        \EndIf
        \EndIf
        \LeftComment{\textbf{\textcolor{black}{Advice Imitation (Ilhan et al., 2021)}}}
        \State $M_\eta, c_2, n_{last}, t_{last} \gets$ 
        $BuildAdvModel(D,M_\eta,n_{last},n_{min},t_{min},t,t_{last},k_{init},
        k_{periodic}$)
        \LeftComment{\textbf{\textcolor{black}{Advice Reuse (Ilhan et al., 2021)}}}
        \State $u_m \gets M_{\eta}^u(s_t)$ \Comment {measure model uncertainty}
        \If{$reuse\_enabled$ is $True$ \textbf{and} $a_t$ is None \textbf{and} $M_{\eta}$ is trained 
        \textbf{and} $u_m < c_2$ } 
        \State $a_t \gets \argmax_{a} M_\eta(a|s_t)$ \Comment {Reuse action}
        \EndIf
        \State Decay $\rho$ w.r.t. pre-defined schedule if $\rho > \rho_{final}$ 
        \LeftComment{\textbf{\textcolor{black}{Self Policy (Ilhan et al., 2021)}}}
        \If {$a_t$ is None}
        \State $a_t \gets \pi_{S}(s_t)$ \Comment{follow student's current policy}
        \EndIf
        \State Take action $a_t$, obtain $s_{t+1},r_{t}$ from $Env$
        \State $D_{dqn}$.append($\left \langle s_t,a_t,r_t,s_{t+1}\right \rangle$) \Comment {store experience}
        \State Update DQN-student-model
        \State Update twin DQN $H_{\omega}$ with DQN's mini-batch \& target
        \State $s_t \gets s_{t+1}$
        \EndFor
        \end{algorithmic}
    \end{algorithm*}
    \clearpage
    
    

\begin{table*}[!pht]
\centering
\begin{tabular}{|l|l|l|}
\hline
\textbf{Hyperparameter}                  & \textbf{Value}         & \textbf{Source or Selected from}                             \\ \hline
Dropout rate                             & 0.2                    & \cite{chen2017agent}                     \\ \hline
No. of forward passes $N$                & 100                    & \cite{DBLP:journals/corr/abs-2104-08440} \\ \hline
Learning rate                            & $6.25 \times 10^{-5} $ & \cite{DBLP:journals/corr/abs-2104-08440} \\ \hline
Minibatch size                           & 32                     & \cite{DBLP:journals/corr/abs-2104-08440} \\ \hline
Discount factor $\gamma$                 & 0.99                   & \cite{DBLP:journals/corr/abs-2104-08440} \\ \hline
Replay memory min. size                  & 10k                    & {(}10k, 50k{)}                                            \\ \hline
Replay memory max. size                  & 500k                   & \cite{DBLP:journals/corr/abs-2104-08440} \\ \hline
Target network update frequency          & 7500                   & \cite{DBLP:journals/corr/abs-2104-08440} \\ \hline
$\epsilon_{initial}$, $\epsilon_{final}$ & 1.0, 0.01              & \cite{DBLP:journals/corr/abs-2104-08440} \\ \hline
total $\epsilon$ decaying steps          & 250k                   & {(}250k, 500k{)}                                          \\ \hline
Percentile $p_1$                         & 70                     & {(}70, 80, 90{)}                                          \\ \hline
$min\_window\_size$ for $D_u$              & 200                    & see text below \\ \hline
Maximum window size for $D_u$            & 10k                    & {(}5k, 10k{)}                                             \\ \hline
Teaching budget $b$                      & 25k                    & {(}12.5k, 25k, 100k{)}                                    \\ \hline
\end{tabular}
\caption{List of all hyperparameters for the student agents. Hyperparameters such as drop out rate, no. of forward passes, percentile, minimum, and maximum uncertainty buffer window size are pertinent to SUA and SUA-AIR. Similarly, teaching budget is applicable to all agents (except NA).}
\label{appendix:hyperparam-students}
\end{table*}


\begin{table*}[!pht]
\centering
\begin{tabular}{|l|l|}
\hline
\textbf{Hyperparameter}                  & \textbf{Value}                                     \\ \hline
Dropout rate                             & 0.35                                    \\ \hline
No. of forward passes               & 100                      \\ \hline
Learning rate                            & $1 \times 10^{-4} $   \\ \hline
Minibatch size                           & 32                       \\ \hline
$\rho_{init}$, $\rho_{final}$, total $\rho$ decaying steps  &  0.1, 0.5 1.5M \\ \hline
Percentile $p_2$                         & 90                      \\ \hline
$t_{min}$, $n_{min}$             & 50k, 2.5k                       \\ \hline
$k_{init}$, $k_{periodic}$          & 50k, 20k                   \\ \hline
\end{tabular}
\caption{List of all hyperparameters for the model of the teacher for AIR and SUA-AIR taken from AIR \cite{DBLP:journals/corr/abs-2104-08440}.}
\label{appendix:hyperparam-model}
\end{table*}
\end{document}